\documentclass[runningheads]{llncs}
\usepackage{graphicx}
\usepackage{url}
\usepackage{amssymb, amsmath, bm}
\usepackage{hyperref}
\usepackage{multirow}
\DeclareMathOperator*{\argmax}{argmax}

\begin{document}
\title{Effective Representation \\for Easy-First Dependency Parsing}

\author{
	Zuchao Li$^{1,2}$,
	Jiaxun Cai$^{1,2}$,
	Hai Zhao$^{1,2,}$\thanks{$\ $ Corresponding author. This paper was partially supported by National Key Research and Development Program of China (No. 2017YFB0304100) and Key Projects of National Natural Science Foundation of China (U1836222 and 61733011).}
}

\authorrunning{Zuchao Li, Jiaxun Cai, Hai Zhao}

\institute{Department of Computer Science and Engineering, Shanghai Jiao Tong University \and
Key Laboratory of Shanghai Education Commission for Intelligent Interaction \\
and Cognitive Engineering, Shanghai Jiao Tong University, Shanghai, China
\email{\{charlee,caijiaxun\}@sjtu.edu.cn, zhaohai@cs.sjtu.edu.cn}}
\maketitle              % typeset the header of the contribution
\begin{abstract}
Easy-first parsing relies on subtree re-ranking to build the complete parse tree. Whereas the intermediate state of parsing processing is represented by various subtrees, whose internal structural information is the key lead for later parsing action decisions, we explore a better representation for such subtrees. In detail, this work introduces a bottom-up subtree encoding method based on the child-sum tree-LSTM. Starting from an easy-first dependency parser without other handcraft features, we show that the effective subtree encoder does promote the parsing process, and can make a greedy search easy-first parser achieve promising results on benchmark treebanks compared to state-of-the-art baselines. Furthermore, with the help of the current pre-training language model, we further improve the state-of-the-art results of the easy-first approach.

\end{abstract}
\section{Introduction}
Transition-based and graph-based parsers are two typical models used in dependency parsing. The former \cite{nivre2003efficient} can adopt rich features in the parsing process but are subject to limited searching space, while the latter \cite{eisner1996efficient,mcdonald2005online,ma2012fourth} searches the entire tree space but limits to local features with higher computational costs. Besides, some other variants are proposed to overcome the shortcomings of both graph and transition based approaches. Easy-first parsing approach \cite{goldberg2010efficient} is introduced by adopting ideas from the both models and is expected to benefit from the nature of the both. Ensemble method \cite{kuncoro-EtAl:2016:EMNLP2016} was also proposed, which employs the parsing result of a parser to guide another in the parsing process. 

Most recent works promote the parsing process by feature refinement. Instead, this work will explore the intermediate feature representation in the incremental easy-first parsing process. Easy-first dependency parser formalizes the parsing process as a sequence of attachments that build the dependency tree bottom-up. Inspired by the fact that humans always parse a natural language sentence starting from the easy and local attachment decisions and proceeding to the harder part instead of working in fixed left-to-right order, the easy-first parser learns its own notion of easy and hard, and defers the attachment decisions it considers to be harder until sufficient information is available. In the primitive easy-first parsing process, each attachment would simply delete the child node and leave the parent node unmodified. However, as the partially built dependency structures always carry rich information to guide the parsing process, effectively encoding those structures at each attachment would hopefully improve the performance of the parser. 

\begin{figure}
	\centering
	\includegraphics[width=0.47\textwidth]{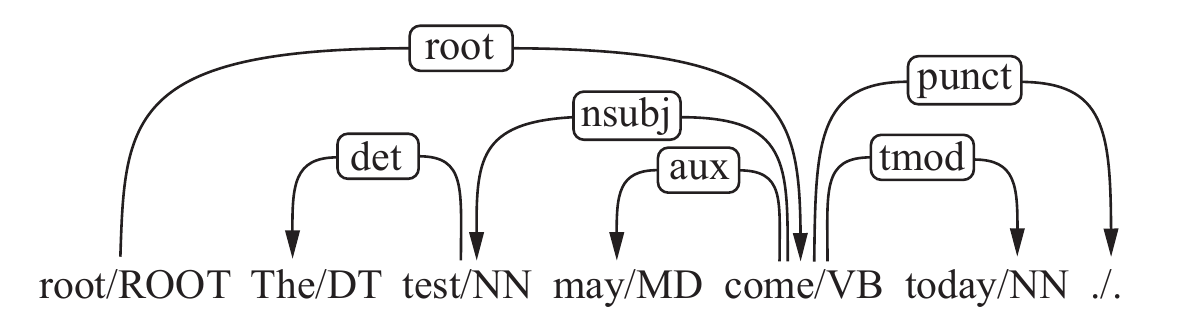}
	\caption{A fully built dependency tree with part-of-speech (POS) tags and \emph{root} token.} \label{fig1}
\end{figure}

There exists a series of studies on encoding the tree structure created in different natural language processing (NLP) tasks using either recurrent neural network or recursive neural network \cite{goller1996learning,socher2010learning}. However, most works require the encoded tree to have fixed maximum factors, and thus are unsuitable for encoding dependency tree where each node could have an arbitrary number of children. Other attempts allow arbitrary branching factors and have succeeded in particular NLP tasks.

\cite{tai-socher-manning:2015:ACL-IJCNLP} introduces a child-sum tree-structured Long Short-Term Memory (LSTM) to encode a completed dependency tree without limitation on branching factors, and shows that the proposed tree-LSTM is effective on semantic relatedness task and sentiment classification task. 
\cite{Zhu2015A} proposes a recursive convolutional neural network (RCNN) architecture to capture syntactic and compositional-semantic representations of phrases and words in a dependency tree and then uses it to re-rank the $k$-best list of candidate dependency trees.
\cite{TACL798} employs two vanilla LSTMs to encode a partially built dependency tree during parsing: one encodes the sequence of left-modifiers from the head outwards, and the other encodes the sequence of right-modifiers in the same manner.

In this paper, we inspect into the bottom-up building process of the easy-first parser and introduce pre-trained language model features and a subtree encoder for more effective representation to promote the parsing process\footnote{Our code is available at \href{https://github.com/bcmi220/erefdp}{https://github.com/bcmi220/erefdp}.}. Unlike the work in \cite{TACL798} that uses two standard LSTMs to encode the dependency subtree in a sequential manner (which we will refer to as HT-LSTM later in the paper), we employ a structural model that provides the flexibility to incorporate and drop an individual child node of the subtree. Further, we introduce a multilayer perceptron between depths of the subtree to encode other underlying structural information like relation and distance between nodes.

From the evaluation results on the benchmark treebanks, the proposed model gives results greatly better than the baseline parser and outperforms the neural easy-first parser presented by \cite{TACL798}. Besides, our greedy bottom-up parser achieves performance comparable to those parsers that use beam search or re-ranking method \cite{Zhu2015A,andor-EtAl:2016:P16-1}.

\section{Easy-First Parsing Algorithm}\label{algorithm}
Easy-first parsing could be considered as a variation of transition-based parsing method, which builds the dependency tree from easy to hard instead of working in a fixed left-to-right order. The parsing process starts by making easy attachment decisions to build several dependency structures, and then proceeds to the harder and harder ones until a well-formed dependency tree is built. During training, the parser learns its own notion of easy and hard, and learns to defer specific kinds of decisions until more information is available \cite{goldberg2010efficient}. 

The main data structure in the easy-first parser is a list of unattached nodes called \emph{pending}. The parser picks a series of actions from the allowed action set, and applies them upon the elements in the \emph{pending} list. The parsing process stops until the \emph{pending} solely contains the root node of the dependency tree. 

At each step, the parser chooses a specific action $\hat{a}$ on position $i$ using a scoring function \emph{score}($\cdot$), which assigns scores to each possible action on each location based on the current state of the parser. Given an intermediate state of parsing process with \emph{pending} $P=\{p_0, p_1, \cdots, p_N\}$, the attachment action is determined by
\begin{equation*}
	\hat{a} = \argmax\limits_{\substack{act \in \mathcal{A},\ 1\leq i \leq N}} \ score(act(i)),
\end{equation*}
where $\mathcal{A}$ denotes the set of the allowed actions, $i$ is the index of the node in the \emph{pending}. Besides distinguishing the correct attachments from the incorrect ones, the scoring function is supposed to assign the ``easiest'' attachment with the highest score, which in fact determines the parsing order of an input sentence. \cite{goldberg2010efficient} employs a linear model for the scorer:
\begin{equation*}
	score(act(i)) = \bm{w}\cdot \bm{\phi}_{act(i)},
\end{equation*}
where $\bm{\phi}_{act(i)}$ is the feature vector of attachment $act(i)$, and $\bm{w}$ is a parameter that can be learned jointly with other components in the model.

\begin{figure} 
  \centering 
  \begin{minipage}[t]{.45\textwidth} 
    \centering 
    \includegraphics[width=0.9\textwidth]{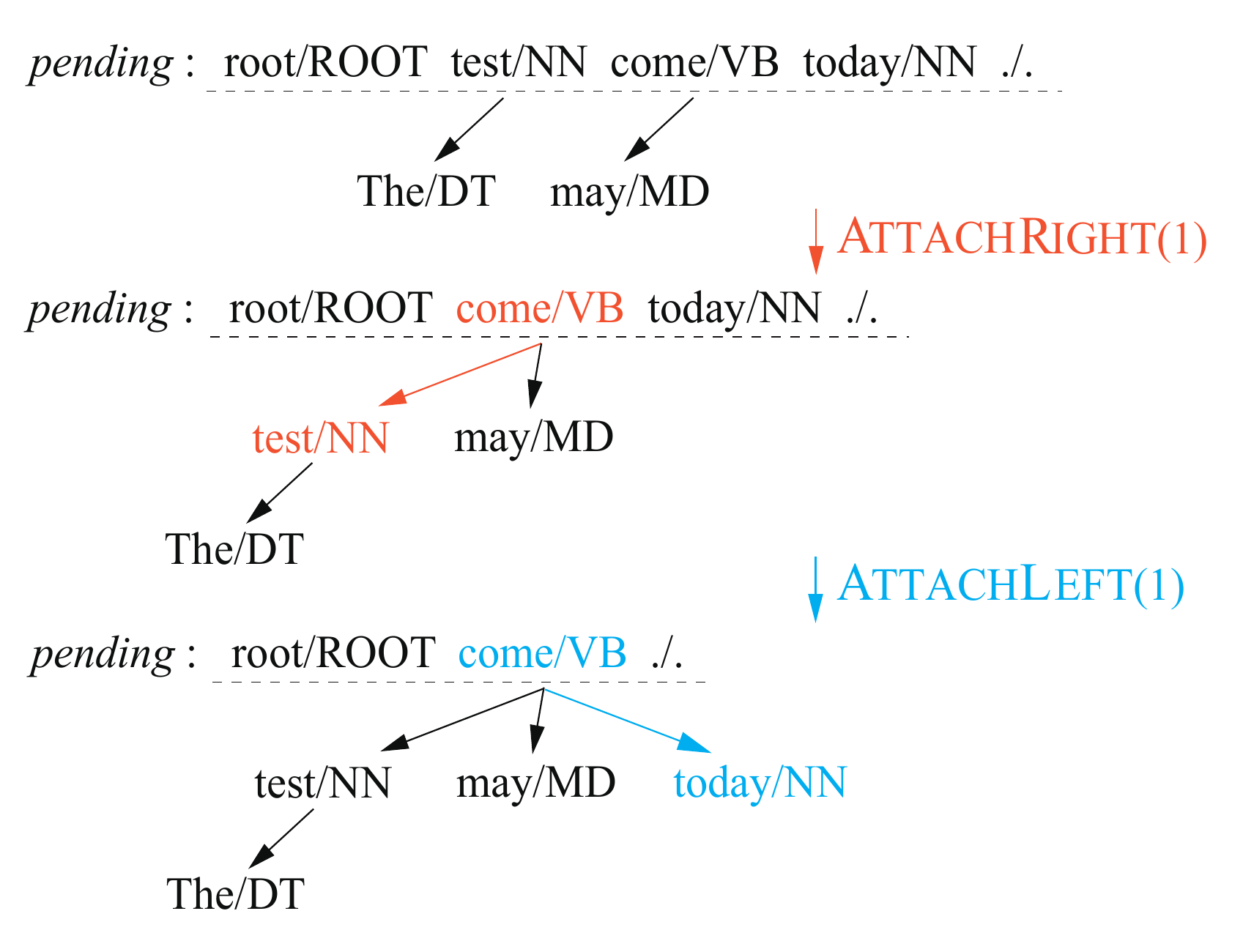}
	\caption{Illustration of the \emph{pending} states before and after the two type of attachment actions}\label{pending_state}
  \end{minipage}% 
  \hspace{.1\textwidth}% 
  \begin{minipage}[t]{.45\textwidth} 
    \centering 
    \includegraphics[width=1.0\textwidth]{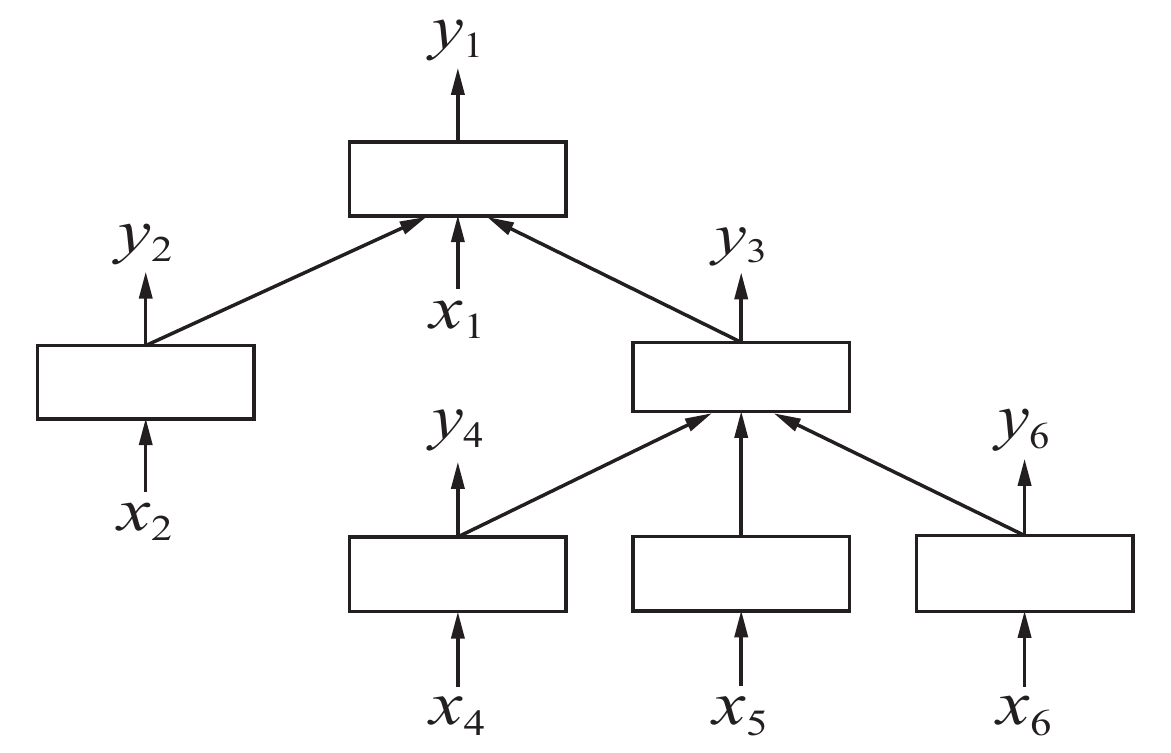}
	\caption{Tree-LSTM neural network with arbitrary number of child nodes}\label{fig4}
  \end{minipage}% 
\end{figure}

There are exactly two types of actions in the allowed action set: \text{\large{A}}TTACH\text{\large{L}}EFT($i$) and \text{\large{A}}TTACH\text{\large{R}}IGHT($i$) as shown in  Figure \ref{pending_state}. Let $p_i$ refer to $i$-th element in the \emph{pending}, then the allowed actions can be formally defined as follows:
\begin{itemize}
	\item \text{\large{A}}TTACH\text{\large{L}}EFT($i$): attaching $p_{i+1}$ to $p_i$ which results in an arc ($p_i$, $p_{i+1}$) headed by $p_i$, and removing $p_{i+1}$ from the \emph{pending}.
	\item \text{\large{A}}TTACH\text{\large{R}}IGHT($i$): attaching $p_i$ to $p_{i+1}$ which results in an arc ($p_{i+1}$, $p_i$) headed by $p_{i+1}$, and removing $p_i$ from the \emph{pending}.
\end{itemize}

%\begin{figure}
%	\centering
%	\includegraphics[width=0.48\textwidth]{pending_states}
%	\caption{Illustration of the \emph{pending} states before and after the two type of attachment actions}\label{pending_state}
%\end{figure}

%\begin{figure}
%	\centering
%	\includegraphics[width=0.4\textwidth]{tree_lstm}
%	\caption{Tree-LSTM neural network with arbitrary number of child nodes}\label{fig4}
%\end{figure}

\section{Parsing With Subtree Encoding}

\subsection{Dependency Subtree}
Easy-first parser builds up a dependency tree incrementally, so in intermediate state, the \emph{pending} of the parser may contain two kinds of nodes: 
\begin{itemize}
	\item \emph{subtree root}: the root of a partially built dependency tree;
	\item \emph{unprocessed node}: the node that has not yet be attached to a parent or assigned a child.
\end{itemize}

Note that each processed node should become a subtree root (attached as a parent) or be removed from the \emph{pending} (attached as a child). A subtree root in the \emph{pending} actually stands for a dependency structure whose internal nodes are all processed, excluding the root itself. Therefore it is supposed to be more informative than the unprocessed nodes to guide the latter attachment decisions. 

In the easy-first parsing process, each \emph{pending} node is attached to its parent only after all its children have been collected. Thus, any structure produced in the parsing process is guaranteed to be a dependency subtree that is consistent with the above definition.

\subsection{Recursive Subtree Encoding}
In the primitive easy-first parsing process, the node that has been removed does not affect the parsing process anymore. Thus the subtree structure in the \emph{pending} is simply represented by the root node. However, motivated by the success of encoding the tree structure properly for other NLP tasks \cite{tai-socher-manning:2015:ACL-IJCNLP,TACL798,kuncoro-EtAl:2017:EACLlong}, we employ the child-sum tree-LSTM to encode the dependency subtree in the hope of further parsing performance improvement. 

\subsubsection{Child-Sum Tree-LSTM}
Child-sum tree-LSTM is an extension of standard LSTM proposed by \cite{tai-socher-manning:2015:ACL-IJCNLP} (hereafter referred to tree-LSTM). Like the standard LSTM unit \cite{hochreiter1997long}, each tree-LSTM unit contains an input gate $i_j$, an output gate $o_j$, a memory cell $c_j$ and a hidden state $h_j$. The major difference between tree-LSTM unit and the standard one is that the memory cell updating and the calculation of gating vectors are depended on multiple child units. As shown in Figure \ref{fig4}, a tree-LSTM unit can be connected to numbers of child units and contains one forget gate for each child. This provides tree-LSTM the flexibility to incorporate or drop the information from each child unit.

Given a dependency tree, let $C(j)$ denote the children set of node $j$, $x_j$ denote the input of node $j$. Tree-LSTM can be formulated as follow \cite{tai-socher-manning:2015:ACL-IJCNLP}:
\begin{align}
	\tilde{\bm{h}}_j &= \sum_{k \in C(j)} \bm{h}_k,\label{eq1}\\
	\bm{i}_j &= \sigma(\bm{W}^{(i)}\bm{x}_j + \bm{U}^{(i)}\tilde{\bm{h}}_j + \bm{b}^{(i)}),\nonumber\\
	\bm{f}_{jk} &= \sigma(\bm{W}^{(f)}\bm{x}_j + \bm{U}^{(f)}\bm{h}_{k} + \bm{b}^{(f)}),\label{eq2}\\ 
	\bm{o}_j &= \sigma(\bm{W}^{(o)}\bm{x}_j + \bm{U}^{(o)}\tilde{\bm{h}}_j + \bm{b}^{(o)}),\nonumber\\
	\bm{u}_j &= \tanh(\bm{W}^{(u)}\bm{x}_j + \bm{U}^{(u)}\tilde{\bm{h}}_j + \bm{b}^{(u)}),\nonumber\\
	\bm{c}_j &= \bm{i}_j \odot \bm{u}_j + \sum_{k \in C(j)} \bm{f}_{jk} \odot \bm{c}_k,\nonumber\\
	\bm{h}_j &= \bm{o}_j \odot \tanh(\bm{c}_j).\nonumber
\end{align}
where $k\in C(j)$, and $\bm{h}_k$ is the hidden state of the $k$-th child node, $\bm{c}_j$ is the memory cell of the head node $j$, and $\bm{h}_j$ is the hidden state of node $j$. Note that in Eq.(\ref{eq2}), a single forget gate $\bm{f}_{jk}$ is computed for each hidden state $\bm{h}_k$.

Our subtree encoder uses tree-LSTM as the basic building block incorporated with the distance and relation label.

\subsubsection{Incorporating Distance and Relation Features} 
Distance embedding is a usual way to encode the distance information. In our model, we use vector $\bm{v}^{(d)}_{h, m_k}$ to represent the relative distance of head word $h$ and its $k$-th modifier $m_k$:
\begin{align*}
	d_{h, m_k}&=index(h) - index(m_k),\\
	\bm{v}^{(d)}_{h, m_k}&=Embed^{(d)}(d_{h, m_k}),
\end{align*}
where $index(\cdot)$ is the index of the word in the original input sentence, and $Embed^{(d)}$ represents the distance embeddings lookup table. 

Similarly, the relation label $\bm{v}^{(rel)}_{h, m_k}$ between head-modifier pair $(h, m_k)$ is encoded as a vector according to the relation embeddings lookup table $Embed^{(r)}$. Both of the two embeddings lookup tables are randomly initialized and learned jointly with other parameters in the neural network.

To incorporate the two features, our subtree encoder introduces an additional feature encoding layer between every connected tree-LSTM unit. Specifically, the two feature embeddings are first concatenated to the hidden state of the corresponding child node. Then we apply an affine transformation on the resulted vector $\bm{g}_k$, and further pass the result through a $\tanh$ activation
\begin{align*}
	\bm{g}_k &= \varphi(\bm{h}_k \oplus \bm{v}^{(d)}_{h, m_k} \oplus \bm{v}^{(rel)}_{h, m_k}), \\
	\varphi(\bm{x}) &= \tanh(\bm{W}^{(\varphi)} \bm{x} + \bm{b}^{(\varphi)})
\end{align*}
where $\bm{W}^{(\varphi)}$ and $\bm{b}^{(\varphi)}$ are learnable parameters. After getting $\bm{g}_k$, it is fed into the next tree-LSTM unit. Therefore, the hidden state of child node $\bm{h}_k$ in Eq.(\ref{eq1}) and (\ref{eq2}) is then replaced by $\bm{g}_k$. 

\subsection{The Bottom-Up Constructing Process}
In our model, a dependency subtree is encoded by performing the tree-LSTM transformation on its root node and computing the vector representation of its children recursively until reaching the leaf nodes. More formally, given a partially built dependency tree rooted at node $h$ with children (modifiers): $h.m_1, h.m_2, h.m_3, \cdots$, which may be roots of some smaller subtree.
%\begin{figure}[!htb]
%	\centering
%	\includegraphics[width=0.2\textwidth]{partial_tree}
%\end{figure}

\noindent Then the tree can be encoded like:
\begin{align}
	\bm{\tau}_h &= f(\bm{\omega}_{h.m_1}, \bm{\omega}_{h.m_2}, \bm{\omega}_{h.m_3}, \bm{x}_j),\label{eq4}\\
	\bm{\omega}_{h.m_k} &= \varphi(\bm{\tau}_{h.m_k}, \bm{v}^{(d)}_{h, m_k}, \bm{v}^{(rel)}_{h, m_k}), k\in\{1, 2, 3, \cdots\}\nonumber
\end{align}
where $f$ is the tree-LSTM transformation, $\varphi$ is the above-mentioned feature encoder, $\bm{\tau}_{h.m_k}$ refers to the vector representation of subtree rooted at node $m_k$, and $\bm{x}_j$ denotes the embedding of the root node word $h$. In practice, $\bm{x}_j$ is always a combination of the word embedding and POS-tag embedding or the output of a bidirectional LSTM. We can see clearly that the representation of a fully parse tree can be computed via a recursive process.

When encountering the leaf nodes, the parser regards them as a subtree without any children and thus sets the initial hidden state and memory cell to a zeros vector respectively:
\begin{equation}\label{eq5}
\bm{\tau}^{(leaf)} = f(\bm{0}, \ x^{(leaf)})
\end{equation}

In the easy-first parsing process, each dependency structure in the \emph{pending} is built incrementally. Namely, the parser builds several dependency subtrees separately and then combines them into some larger subtrees. So, when the parser builds a subtree rooted at $h$, all its children have been processed by some previous steps. The subtree encoding process can be naturally incorporated into the easy-first parsing process in a bottom-up manner using the dynamic programming technique.

Specifically, in the initial step, each node $w_i$ in the input sentence is treated like a subtree without any children. The parser initializes the \emph{pending} with the tree representation $\bm\tau_{w_i}^{(leaf)}$ of those input nodes using Eq.(\ref{eq5}). For each node in \emph{pending}, the parser maintains an additional \emph{children} set to hold their processed children. Each time the parser performs an attachment on the nodes in the \emph{pending}, the selected modifier is removed from \emph{pending} and then added to the children set of the selected head. The vector representation of the subtree rooted at the selected head is recomputed using Eq.(\ref{eq4}). The number of times that the easy-first parser performs updates on the subtree representations is equal to the number of actions required to build a dependency tree, namely, $N$-1, where $N$ is the input sentence length.

\subsection{Incorporating HT-LSTM and RCNN}
Both HT-LSTM and RCNN can be incorporated into our framework. However, since the RCNN model employs POS tag dependent parameters, its primitive form is incompatible with the incremental easy-first parser, for which we leave a detail discussion in appendix \ref{relatedwork}. To address this problem, we simplify and reformulate the RCNN model by replacing the POS tag dependent parameters with a global one. 
Specifically, for each head-modifier pair $(h, m_k)$, we first use a convolutional hidden layer to compute the combination representation:
\begin{equation*}
\begin{split}
	\bm{z_k} &= \tanh(\bm{W}^{(global)}\bm{p}_k), 0 < k \leq K, \\ 
	\bm{p}_k &= \bm{x}_h \oplus \bm{g}_k,\\
	\bm{g}_k &= \varphi(\bm{\tau}_{k} \oplus \bm{v}^{(d)}_{h, m_k} \oplus \bm{v}^{(rel)}_{h, m_k}), \\
\end{split}
\end{equation*}
where $K$ is the size of the children set $C(h)$ of node $h$, $\bm{W}^{(global)}$ is the global composition matrix, $\bm{\tau}_{k}$ is the subtree representation of the child node $m_k$, which can be recursively computed using the RCNN transformation. After convolution, we stack all $\bm{z}_k$ into a matrix $\bm{Z}^{(h)}$. Then to get the subtree representation for $h$, we apply a max pooling over $\bm{Z}^{(h)}$ on rows:
\begin{equation*}
	\bm{\tau}_h = \max_{k}\bm{Z}^{(h)}_{j, k}, 0 < j \leq d, 0 < k \leq K, 
\end{equation*}
where $d$ is the dimensionality of $z_k$.

\begin{table} 
  \centering 
  \begin{minipage}[t]{.45\textwidth} 
    \centering
     \scalebox{0.9}{
     \begin{tabular}{l|cc|cc}
     \hline
     \hline
      & \textbf{Dev}&(\%) & \textbf{Test}&(\%) \\
      & LAS & UAS & LAS & UAS \\
      \hline
      \hline
      BiLSTM parser & 90.73 & 92.87 & 90.67 & 92.83 \\
      \hline
      RCNN & 91.05 & 93.25 & 91.01 & 93.21 \\
      HT-LSTM & 91.23 & 93.23 & 91.36 & 93.27 \\
      tree-LSTM & \textbf{92.32} & \textbf{94.27} & \textbf{92.33} & \textbf{94.31} \\
      \hline
      tree-LSTM + ELMo & \textbf{92.97} & \textbf{94.95} & \textbf{93.09} & \textbf{95.33} \\
      tree-LSTM + BERT & \textbf{93.14} & \textbf{95.68} & \textbf{93.27} & \textbf{95.71} \\
       tree-LSTM + ELMo & \multirow{2}{*}{\textbf{93.44}} & \multirow{2}{*}{\textbf{95.87}} & \multirow{2}{*}{\textbf{93.49}} & \multirow{2}{*}{\textbf{95.87}}\\
       + BERT & & \\
      \hline
      \hline
      \end{tabular}
      }
     \caption{Comparison with baseline easy-first parser.} \label{ablation}
  \end{minipage}% 
  \hspace{.1\textwidth}% 
  \begin{minipage}[t]{.4\textwidth} 
    \centering
     \scalebox{0.9}{
     \begin{tabular}{l|cc|cc}
     \hline
     \hline
      & \textbf{Dev}&(\%) & \textbf{Test}&(\%) \\
      & LAS & UAS & LAS & UAS \\
      \hline
      \hline
      baseline parser & 78.83 & 82.97 & 78.43 & 82.55 \\
      \hline
      +tree-LSTM & 91.10 & 92.98 & 91.08 & 92.94 \\
      \ \ +Bi-LSTM & 91.62 & 93.49 & 91.52 & 93.46 \\
      \ \ \ \ +pre-train & 92.01 & 93.97 & 91.95 & 93.95 \\
      \hline
      \hline
      baseline parser$^{*}$ & 79.0 & 83.3 & 78.6 & 82.7 \\
      \hline
      +HT-LSTM$^{*}$ & 90.1 & 92.4 & 89.8 & 92.0 \\
      \ \ +Bi-LSTM$^{*}$ & 90.5 & 93.0 & 90.2 & 92.6 \\
      \ \ \ \ +pre-train $^{*}$ & 90.8 & 93.3 & 90.9 & 93.0 \\
      \hline
      \hline
     \end{tabular}
     }
     \caption{Results under the same settings reported in \protect\cite{TACL798}.} \label{ht-ablation} %hyper-parameters 
  \end{minipage}% 
\end{table}

% \begin{table}
%  \centering
%  \scalebox{0.9}{
%  \begin{tabular}{l|cc|cc}
%  \hline
%  \hline
%   & \textbf{Dev}&(\%) & \textbf{Test}&(\%) \\
%   & LAS & UAS & LAS & UAS \\
%   \hline
%   \hline
%   BiLSTM parser & 90.73 & 92.87 & 90.67 & 92.83 \\
%   \hline
%   RCNN & 91.05 & 93.25 & 91.01 & 93.21 \\
%   HT-LSTM & 91.23 & 93.23 & 91.36 & 93.27 \\
%   tree-LSTM & \textbf{92.32} & \textbf{94.27} & \textbf{92.33} & \textbf{94.31} \\
%   \hline
%   tree-LSTM + ELMo & \textbf{92.97} & \textbf{94.95} & \textbf{93.09} & \textbf{95.33} \\
%   tree-LSTM + BERT & \textbf{93.14} & \textbf{95.68} & \textbf{93.27} & \textbf{95.71} \\
%   tree-LSTM + ELMo + BERT & \textbf{93.44} & \textbf{95.87} & \textbf{93.49} & \textbf{95.87}\\
%   \hline
%   \hline
%   \end{tabular}
%   }
%  \caption{Comparison with baseline easy-first parser.} \label{ablation}
% \end{table}

\section{Experiments and Results}
We evaluate our parsing model on English Penn Treebank (PTB) and Chinese Penn Treebank (CTB), using unlabeled attachment scores (UAS) and labeled attachment scores (LAS) as the metrics. Punctuations are ignored as in previous work \cite{TACL798,dozat2017deep}. Pre-trained word embeddings and language model have been shown useful in a lot of tasks. Therefore, we also add the latest ELMo \cite{peters2018deep} and BERT \cite{devlin2018bert} pre-trained language model layer features to enhance our representation. 
%The training details are shown in the appendix \ref{training_detail}.

\subsection{Treebanks}
For English, we use the Stanford Dependency (SD 3.3.0) \cite{de2008stanford} conversion of the Penn Treebank \cite{marcus1993building}, and follow the standard splitting convention for PTB, using sections 2-21 for training, section 22 as development set and section 23 as test set. Stanford POS tagger \cite{toutanova2003feature} is to give predicted POS tags.

For Chinese, we adopt the splitting convention for CTB described in \cite{zhao2008parsing,zhao2009cross,zhang2016probabilistic,dyer2015transition}. The dependencies are converted with the Penn2Malt converter. Gold segmentation and POS tags are used as in previous work \cite{dyer2015transition}.

\subsection{Results}
\subsubsection{Improvement over Baseline Model}

To explore the effectiveness of the proposed subtree encoding model, we implement a baseline easy-first parser without additional subtree encoders and conduct experiments on PTB. The baseline model contains a BiLSTM encoder and uses pre-trained word embedding, which we refer to BiLSTM parser. We also re-implement both HT-LSTM and RCNN and incorporate them into our framework for subtree encoding. All the four models share the same hyper-parameters settings and the same neural components except the subtree encoder. 

The results in Table \ref{ablation} show that our proposed tree-LSTM encoder model outperforms the BiLSTM parser with a margin of 1.48\% in UAS and 1.66\% in LAS on the test set. Though the RCNN model keeps simple by just using a single global matrix $\bm{W}^{(global)}$, it draws with the HT-LSTM model in UAS on both the development set and the test set, and slightly underperforms the latter one in LAS. Note that the HT-LSTM is more complicated, which contains two LSTMs. Such results demonstrate that simply sequentializing the subtree fails to effectively incorporate the structural information. A further error analysis of the three models is given in the following section.

Besides, to make a fair comparison, we also run our model under the same setting as those reported in \cite{TACL798}, and report the results in Table \ref{ht-ablation}. Experiment results show that the performance of the tree-LSTM parser declines slightly but still outperforms the HT-LSTM parser. The ``+'' symbol denotes a specific extension over the previous line. The results with $^{*}$ is reported in \protect\cite{TACL798}. It is worth noting that their weak baseline parser does not use Bi-LSTM and pre-trained embeddings.

\begin{table}
 \centering
 \scalebox{0.9}{
 \begin{tabular}{l|c|cc|cc}
 \hline
 \hline
   & &\textbf{PTB-SD}& & \textbf{CTB}& \\
  \textbf{System} & \textbf{Method} & LAS(\%) & UAS(\%) & LAS(\%) & UAS(\%) \\
  \hline
  \hline
  (Dyer et al., 2015) \cite{dyer2015transition} & Transition (g) & 90.9 & 93.1 & 85.5 & 87.1 \\
  (Kiperwasser and Goldberg, 2016b) \cite{TACL885} & Transition (g) & 91.9 & 93.9 & \textbf{86.1} & \textbf{87.6} \\
  (Andor et al., 2016) \cite{andor-EtAl:2016:P16-1} & Transition (b)  & \textbf{92.79} & \textbf{94.61} & - & - \\
  (Zhu et al., 2015) \cite{Zhu2015A} & Transition (re) & - & 94.16 & - & 87.43 \\
  \hline
  (Zhang and McDonald, 2014) \cite{zhang-mcdonald:2014:P14-2} & Graph (3rd)& 90.64 & 93.01 & 86.34 & 87.96 \\
  (Wang and Chang, 2016) \cite{wang-chang:2016:P16-1} & Graph (1st)& 91.82 & 94.08 & 86.23 & 87.55 \\
  (Kiperwasser and Goldberg, 2016b) \cite{TACL885} & Graph (1st)& 90.9 & 93.0 & 84.9 & 86.5 \\
  (Dozat and Manning, 2017) \cite{dozat2017deep} & Graph (1st)& \textbf{94.08} & \textbf{95.74} & \textbf{88.23} & \textbf{89.30} \\
  (Wang et al., 2018) & Graph (1st)& \textbf{94.54} & 95.66 & - & - \\
  (Wang et al., 2018) \cite{wang2018improved} + ELMo & Graph (1st)& \textbf{95.25} & \textbf{96.35} & - & - \\
  \hline
  (Zhang et al., 2017) \cite{zhang2017stack} & Seq2seq (b) & 91.60 & 93.71 & 85.40 & 87.41 \\
  (Li et al., 2018) \cite{li2018seq2seq}& Seq2seq (b) & 92.08 & 94.11 & 86.23 & 88.78\\
  \hline
  (Kiperwasser and Goldberg, 2016a) \cite{TACL798} & EasyFirst (g)  & 90.9 & 93.0 & 85.5 & 87.1 \\
  \textbf{This work} & EasyFirst (g) & \textbf{92.33} & \textbf{94.31} & \textbf{86.37} & \textbf{88.65} \\
  \hline
  \textbf{This work + ELMo} & EasyFirst (g) & \textbf{93.09} & \textbf{95.33} & - & -\\
  \textbf{This work + BERT} & EasyFirst (g) & \textbf{93.27} & \textbf{95.71} & \textbf{87.44} & \textbf{89.52} \\
  \textbf{This work + ELMo + BERT} & EasyFirst (g) & \textbf{93.49} & \textbf{95.87} & - & - \\
 \hline
 \hline
 \end{tabular}
 }
 \caption{Comparison of results on the test sets. Acronyms used: (g) -- greedy, (b) -- beam search, (re) -- re-ranking, (3rd) -- 3rd-order, (1st) -- 1st-order. Because ELMo does not have a Chinese version, the ``+ELMo" rows have no results.} \label{overall}
\end{table}

\subsubsection{Comparison with Previous Parsers}
We now compare our model with some other recently proposed parsers. The results are compared in Table \ref{overall}. 
%It is worth noting that although the work in \cite{kuncoro-EtAl:2017:EACLlong} can reach an accuracy of 95.8\% (UAS) on PTB, it is not included in the table since its parsing result is converted from phrase-structure parsing, while this work focuses on the native dependency parsing method. 
The work in \cite{TACL798} (HT-LSTM) is similar to ours and achieves the best result among the recently proposed easy-first parsers\footnote{Here we directly refer to the original results reported in \protect\cite{TACL798}}. Our subtree encoding parser outperforms their model on both PTB and CTB. Besides, the proposed model also outperforms the RCNN based re-ranking model in \cite{Zhu2015A}, which introduces an RCNN to encode the dependency tree and re-ranks the $k$-best trees produced by the base model. Note that although our model is based on the greedy easy-first parsing algorithm, it is also competitive to the search-based parser in \cite{andor-EtAl:2016:P16-1}. The model in \cite{dozat2017deep} outperforms ours, however, their parser is graph-based and thus can enjoy the benefits of global optimization.

\begin{figure} 
  \centering 
  \begin{minipage}[t]{.45\textwidth} 
    \centering 
    \includegraphics[width=1.0\textwidth]{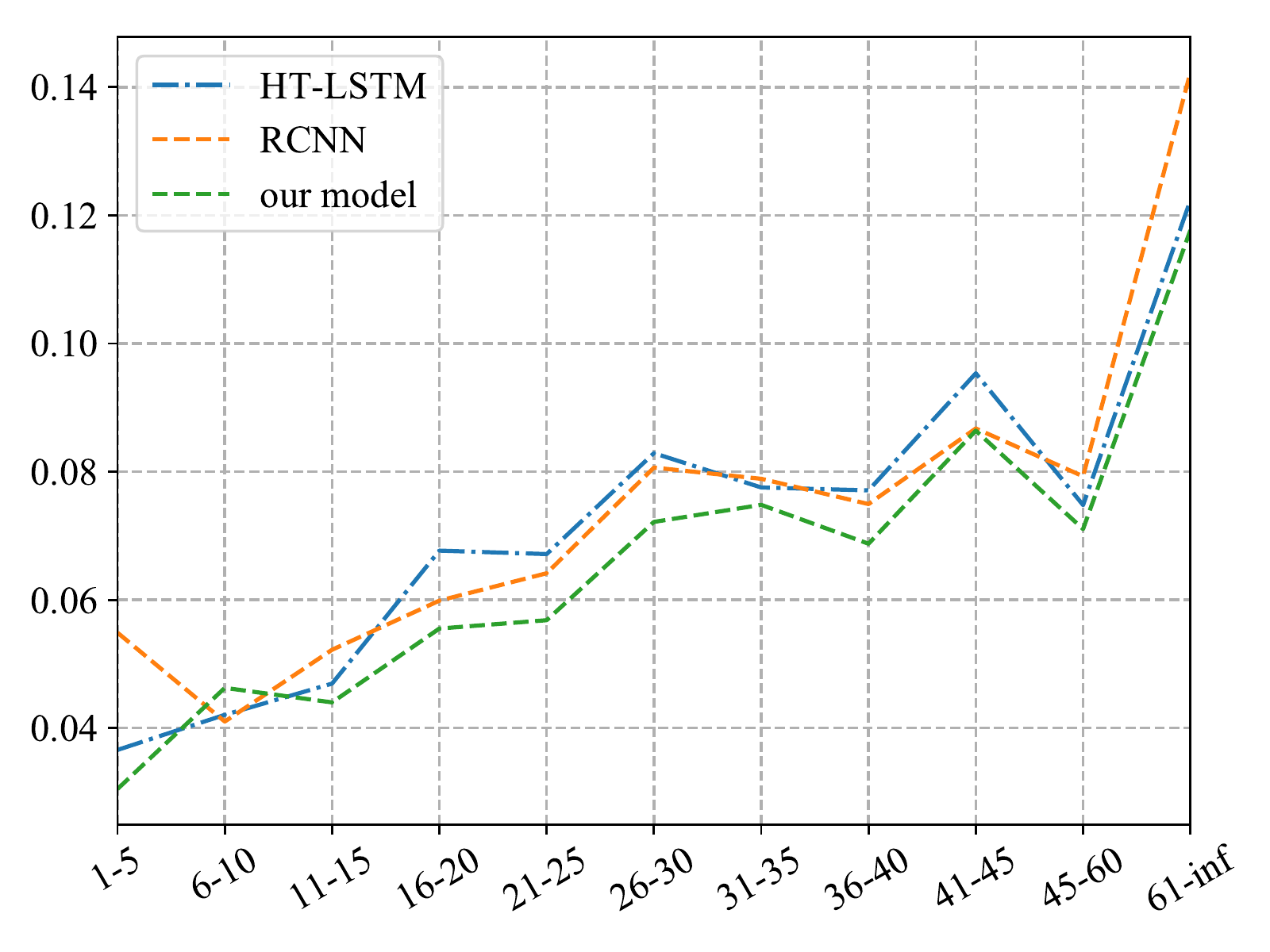}
	\caption{Line chart of error rate against sentence length} \label{length}
  \end{minipage}% 
  \hspace{.1\textwidth}% 
  \begin{minipage}[t]{.45\textwidth} 
    \centering 
    \includegraphics[width=0.9\textwidth]{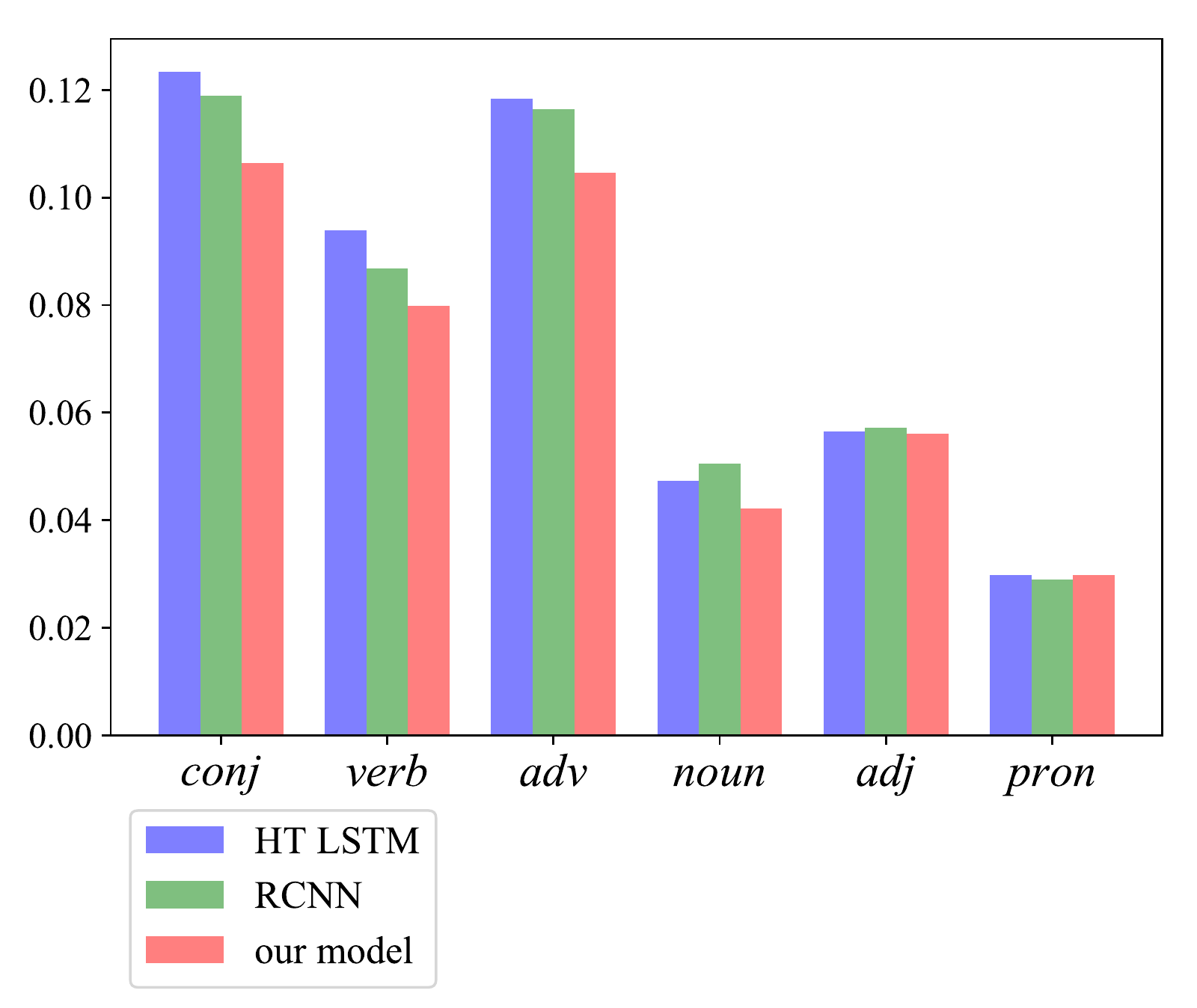}
	\caption{Error rate with respect to POS tags} \label{postags}
  \end{minipage}% 
\end{figure}

\subsection{Error Analysis}\label{analysis}
To characterize the errors made by parsers and the performance enhancement by importing the subtree encoder, we present some analysis on the error rate with respect to the sentence length and POS tags. All analysis is conducted on the unlabeled attachment results from the PTB development set.

\subsubsection{Error Distribution over Dependency Distance}
%It is well known that the dependency parsers are not good at coping with long-distance dependency. 
Figure \ref{length} shows the error rate of different subtree encoding methods with respect to sentence length. 
The error rate curves of the three models share the same tendency: as the sentence length grows, the error rate increases. In most of the cases, the curve of our model lies below the other two curves, except the case that the sentence length lies in $6$-$10$ where the proposed model underperforms the other two with a margin smaller than 1\%. The curve of HT-LSTM and that of RCNN cross with each other at several points. It is not surprising since the overall results of the two models are very close. The curves further show that tree-LSTM is more suitable for incorporating the structural information carried by the subtrees produced in the easy-first parsing process.

%\begin{figure}
%	\centering
%	\includegraphics[width=0.47\textwidth]{against_length}
%	\caption{Line chart of error rate against sentence length} \label{length}
%\end{figure}

%\begin{figure}
%	\centering
%	\includegraphics[width=0.47\textwidth]{postags}
%	\caption{Error rate with respect to POS tags} \label{postags}
%\end{figure}

\subsubsection{Error Distribution over POS tags}
\cite{mcdonald-nivre:2007:EMNLP-CoNLL2007} distinguishes \emph{noun}, \emph{verb}, \emph{pronoun}, \emph{adjective}, \emph{adverb}, \emph{conjunction} for POS tags to perform a linguistic factors analysis. To follow their works, we conduct a mapping on the PTB POS tags and skip those which cannot be mapped into one of the six above-mentioned POS tags. Then we evaluate the error rate with respected to the mapped POS tags and compare the performance of the three parsers in Figure \ref{postags}. 

The results seem to be contradicted with the previous ones at first sight since the HT-LSTM model underperforms the RCNN one in most cases. This interesting result is caused by the overwhelming number of \emph{noun}. According to statistics, the number of \emph{noun} is roughly equal to the total number of \emph{verb}, \emph{adverb} and \emph{conjunction}. 

Typically, the \emph{verb}, \emph{conjunction} and \emph{adverb} tend to be closer to the root in a parse tree, which leads to a longer-distance dependency and makes it more difficult to parse. The figure shows that our model copes better with those kinds of words than the other two models. %Interestingly, the simple RCNN model outperforms the HT-LSTM model on all three categories of words, which demonstrates that the HT-LSTM hardly succeeds to capture long-distance dependency through the tree structure. This might be caused by the sequentialization of the subtree in the HT-LSTM model.   

The other three categories of words are always attached lower in a parse tree and theoretically should be easier to parse. In the result, the three models perform similarly on \emph{adjective} and \emph{pronoun}. However, the RCNN model performs worse than the other two models on \emph{noun}, which can be attributed to too simple RCNN model that is unable to cover different lengths of dependency. 

\subsection{Related Work}\label{relatedwork}
Recently, neural networks have been adopted for a wide range of traditional NLP tasks \cite{zhao:sighan5,zhao:sighan6,zhao:paclic20,zhao2009multilingual,zhao:talip2010,zhao:info2011,zhao:jair2013}. A recent line of studies including Chinese word segmentation \cite{cd:acl2016,cd:acl2017}, syntactic parsing \cite{zhang2016probabilistic,li2018joint,li2018seq2seq}, semantic role labeling \cite{he2018syntax,cai2018full,li2018unified} and other NLP applications \cite{zhang2018DUA,huang2018moon,xiao2018prediction,zhang2018exploring,AAAI1816060} have drawn a lot of attention. Easy-first parsing has a special position in dependency parsing system. As mentioned above, to some extent, it is a kind of hybrid model that shares features with both transition and graph based models, though quite a lot of researchers still agree that it belongs to the transition-based type as it still builds parse tree step by step. Since easy-first parser was first proposed in \cite{goldberg2010efficient}, the most progress on this type of parsers is \cite{TACL798} who incorporated neural network for the first time.
% Easy-first parsing has a special position in dependency parsing system, and it was first proposed in \cite{goldberg2010efficient}, the most progress on this type of parsers is \cite{TACL798} who incorporated the neural network for the first time.

Most of the RNNs are limited to a fixed maximum number of factors \cite{socher2014recursive}. To release the constraint of the limitation of factors, \cite{Zhu2015A} augments the RNN with a convolutional layer, resulting in a recursive convolutional neural network (RCNN). The RCNN is able to encode a tree structure with an arbitrary number of factors, and is used in a re-ranking model for dependency parsing. %The primitive RCNN employs POS tag dependent parameters which prevent it from being conveniently incorporated in an incremental parser which causes a data sparsity problem. The great number of parameters and imbalance of training data would make the model hard to train.

Child-sum tree-LSTM \cite{tai-socher-manning:2015:ACL-IJCNLP} is a variant of the standard LSTM which is capable of getting rid of the arity restriction, and has been shown effective on semantic relatedness and the sentiment classification tasks. We adopt the child-sum tree-LSTM in our incremental easy-first parser to promote the parsing.

% The work in \cite{TACL798} is similar to ours, which sequentializes the dependency subtree and then encodes the modifiers from the head outward. As HT-LSTM models the modifiers of a head as an ordered sequence, each time the subtree grows with a new adding modifier, its choice heavily depends on the previous one. If the parser makes an incorrect attachment on its $j$-th right modifier, then the error will propagate to all right modifiers after $j$ through the vanilla LSTM. By contrast, in our subtree encoder model, since each modifier is assigned with a single forget gate, the encoder is able to drop an individual attachment, which can effectively alleviate the problem of error propagation.

\section{Conclusion} % 
To enhance the easy-first dependency parsing, this paper proposes a tree encoder and integrates pre-trained language model features for a better representation of partially built dependency subtrees. Experiments on PTB and CTB verify the effectiveness of the proposed model. %As easy-first parser builds dependency tree bottom-up and stops when there is only one node (the ``ROOT") remained in the pending, our subtree encoder would get a representation for the full built dependent tree once the parsing process is done. The last tree representation can be directly integrated into some other downstream application. One of our future works is to verify the effectiveness of our subtree encoder by applying it on some other tasks like textual inference or even using the subtree representation as an additional feature for neural machine translation. 

\bibliographystyle{splncs04}
\bibliography{references}

\end{document}